\def\year{2022}\relax
\documentclass[letterpaper]{article} 
\usepackage{aaai22}  
\usepackage{times}  
\usepackage{helvet}  
\usepackage{courier}  
\usepackage[hyphens]{url}  
\usepackage{graphicx} 
\usepackage{natbib}  
\usepackage{caption} 
\DeclareCaptionStyle{ruled}{labelfont=normalfont,labelsep=colon,strut=off} 
\usepackage{algorithm}
\usepackage{algorithmic}

\usepackage{epsfig}
\usepackage{microtype}
\graphicspath{{./figure/}}
\usepackage{subfigure}
\usepackage{multirow}
\usepackage{enumitem}
\usepackage{amsmath}
\usepackage{amssymb}
\usepackage[dvipsnames]{xcolor}

\usepackage{newfloat}
\usepackage{listings}
\floatstyle{ruled}
\newfloat{listing}{tb}{lst}{}
\floatname{listing}{Listing}

\title{TRACER: Extreme Attention Guided Salient Object Tracing Network}
\author {
    Min Seok Lee\textsuperscript{\rm 1}\equalcontrib,
    Wooseok Shin\textsuperscript{\rm 1}\equalcontrib,
    Sung Won Han\textsuperscript{\rm 1}\thanks{Corresponding author.}
}
\affiliations {
    \textsuperscript{\rm 1} School of Industrial and Management Engineering, Korea University\\
    \{karel, wsshin95, swhan\}@korea.ac.kr
}

\begin{document}

\maketitle

\begin{abstract}
Existing studies on salient object detection (SOD) focus on extracting distinct objects with edge information and aggregating multi-level features to improve SOD performance. To achieve satisfactory performance, the methods employ refined edge information and low multi-level discrepancy. However, both performance gain and computational efficiency cannot be attained, which has motivated us to study the inefficiencies in existing encoder-decoder structures to avoid this trade-off. We propose TRACER, which detects salient objects with explicit edges by incorporating attention guided tracing modules. We employ a masked edge attention module at the end of the first encoder using a fast Fourier transform to propagate the refined edge information to the downstream feature extraction. In the multi-level aggregation phase, the union attention module identifies the complementary channel and important spatial information. To improve the decoder performance and computational efficiency, we minimize the decoder block usage with object attention module. This module extracts undetected objects and edge information from refined channels and spatial representations. Subsequently, we propose an adaptive pixel intensity loss function to deal with the relatively important pixels unlike conventional loss functions which treat all pixels equally. A comparison with 13 existing methods reveals that TRACER achieves state-of-the-art performance on five benchmark datasets. We have released TRACER at \url{https://github.com/Karel911/TRACER}.
\end{abstract}


\section{Introduction}
To improve the performance of salient object detection (SOD), existing methods can be categorized into two approaches: those that improve the edge representation and those that reduce discrepancies during multi-level aggregation. To refine edges, existing methods incorporated shallow encoder representations, which contain sufficient edge information \cite{zhang2017amulet, wang2018detect, wu2019cascaded}, and deeper encoder outputs \cite{feng2019attentive, qin2019basnet, wu2019stacked, zhao2019egnet, wei2020label}. Although multi-level aggregation can explicitly represent edges, multiple encoder outputs have different data distributions. For the discrepancy reduction, skip-connected or cascaded decoder structures have been proposed \cite{zhang2018exfuse, zhou2018unet++, feng2019attentive, pang2019towards, qin2019basnet, zhao2019egnet}. These existing approaches improved SOD performance; however, they are incapable of simultaneously achieving the performance and computational efficiency. Therefore, to improve both performance and computational efficiency, this study focuses on reducing inefficiencies, which can develop in existing encoder-decoder structures, and applying adaptive pixel-wise weights to conventional loss functions.

In existing encoder-decoder structures, previous studies used multi-level encoder representations across a network \cite{feng2019attentive, qin2019basnet, wu2019stacked, zhao2019egnet, wei2020label} to extract edge information. Although these studies improve SOD performance because of the explicit edge information, they extensively employ edge refinement modules, which require all or part of the encoder outputs. These approaches increase parameter overhead and computational inefficiency. Moreover, the methods in these studies do not leverage the refined edges in the downstream feature extraction phases because the edge generating approaches depend on other encoder outputs. Because low- and high-level information can be effective for detecting edge features and semantic representations, the use of the refinement modules at both levels should be determined selectively and minimized for network efficiency.

The purpose of the decoder is to minimize discrepancies during multi-level aggregation when incorporating low- and high-level representations \cite{zhang2018exfuse, pang2019towards, qin2019basnet, wei2020f3net}. Because a decoder that uses multiple encoder outputs consumes a large amount of memory, recent studies have proposed a cascaded decoder, which reduces the low-level connections or aggregates high-level representations \cite{wu2019cascaded, wei2020f3net, wei2020label}. Subsequently, others have proposed object and edge refinement modules \cite{kuen2016recurrent, chen2018reverse, liu2018picanet, zhang2018progressive, qin2019basnet, wang2019salient, zhao2019egnet, zhao2019pyramid} or multi-decoder structures \cite{wu2019stacked, wei2020f3net, wei2020label} to reduce the discrepancies in different representations. However, it is yet to be ascertained whether the multiple modules and multi-decoder are efficient for computation. To improve performance and network efficiency, the focus should be on determining which representations over multiple levels are important while minimizing the use of attention modules in the multi-level aggregation and decoder structure. Because the different levels contain different data distributions, the distinct representations should be emphasized during multi-level aggregation.

In the process of applying adaptive pixel-wise weights to the loss function, each pixel is treated independently with binary cross entropy (BCE) and IoU losses, which are globally adopted for loss functions \cite{zhao2019egnet, pang2020multi}. It is evident that the pixels adjacent to fine or explicit edges should be more focused than the pixels in the background or center of the salient object. To treat this pixel inconsistency, existing study \cite{wei2020f3net} proposed the weighted BCE and weighted IoU loss functions. Although the study allocated larger weights to relatively important pixels, they also imposed significance on redundant pixels, including background regions while not covering explicit and fine edges. Consequently, it is necessary to employ adaptive pixel-wise weights to delineate fine or explicit edge regions while excluding redundant areas.

This study proposes an extreme attention guided salient object tracing network called TRACER. To address the inefficiencies in existing approaches, we apply three attention guided modules (i.e., masked edge, union, and object attention modules) in the shallow encoder, multi-level aggregation process, and decoder, respectively. The masked edge attention module enhances the edge features in low-level representations using a fast Fourier transform and propagates the edge-refined representation to the next encoder. The union attention module aggregates multi-level encoder outputs to reduce discrepancies in the distributions. Subsequently, this module determines the more important context in the aggregated channel- and spatial-wise representations. Following the integration, the object attention module incorporates low-level encoder representations and decoder outputs to identify the salient objects. To deal with the relative significance of pixels, we propose an adaptive pixel intensity loss function. We aggregate adjacent pixels around the target pixel by employing multiple kernel aggregation and exclude weights outside the edges. When the target pixel consists of fine or explicit edges, it is assigned a higher intensity than the other pixels. The main contributions of this study are as follows.
\begin{itemize}[leftmargin=\dimexpr\parindent+0.1mm+0.1\labelwidth\relax]
    \item We study the inefficiencies of the existing encoder-decoder structures and propose an efficient network called TRACER, which overcomes the trade-off between SOD performance improvement and computational efficiency.
    \item We propose masked edge, union, and object attention modules, which effectively and efficiently identify salient objects and edges in encoder-decoder structures with minimal use.
    \item The adaptive pixel intensity loss function focuses on the relatively significant pixels, which are adjacent to explicit or fine edges. It enables the network to optimize noisy label robustness and local-global structure awareness.
    \item TRACER significantly outperforms the 13 existing methods on five benchmark datasets under four evaluation metrics and achieves computational efficiency.
 \end{itemize}


\section{Related Work}
\subsection{Network performance and efficiency}
Since the introduction of FCN \cite{long2015fully} and U-Net \cite{ronneberger2015u}, existing studies have used VGG \cite{zhang2017amulet, feng2019attentive, wang2019salient, wu2019mutual, zhao2019egnet} and ResNet \cite{qin2019basnet, wu2019stacked, zhao2019egnet, wei2020f3net, wei2020label, pang2020multi} as backbone encoders in the U-shaped encoder-decoder structure. Using the VGG \cite{simonyan2014very} and ResNet \cite{he2016deep}, which have satisfactory generalization performance, existing methods employed various object refinement modules and multi-decoder structures to improve SOD performance. The modules extracted explicit edge information \cite{feng2019attentive, qin2019basnet, wei2020label, wu2019stacked, zhao2019egnet} or detected elaborate object regions to enhance salient object representations \cite{chen2018reverse, pang2020multi, wu2019mutual}. In terms of decoder efficiency, an existing study proposed a cascaded partial decoder, which aggregates deeper encoder representations and excludes shallow outputs to reduce computational inefficiency \cite{wu2019cascaded}. Furthermore, recent studies have improved SOD performance by applying object and edge refinement modules with multi-decoder structures \cite{wu2019stacked, wei2020f3net, wei2020label}. Although these studies help avoid increases in the parameter overhead despite the use of multiple decoders, few approaches focus on improving computational efficiency and performance simultaneously.

\subsection{Edge refinement module efficiency}
Enhancing the edge representation, which improves the detection of a salient object, can lead to a substantial improvement in SOD performance. To represent edge information, existing methods aggregated multi-level representations or applied edge refinement modules in a multi-decoder structure \cite{feng2019attentive, qin2019basnet, wu2019stacked, zhao2019egnet, wei2020label}. However, using multi-level representations or refinement modules with multi-decoders is inefficient with respect to memory and computation. Moreover, previous approaches could not use edges to enhance the encoder outputs in the feature extraction phase because these methods required deeper encoder outputs to obtain distinct edges. To generate edges efficiently, minimizing the use of multiple outputs is necessary.

\begin{figure*}[ht]
\begin{center}
\includegraphics[scale=0.36]{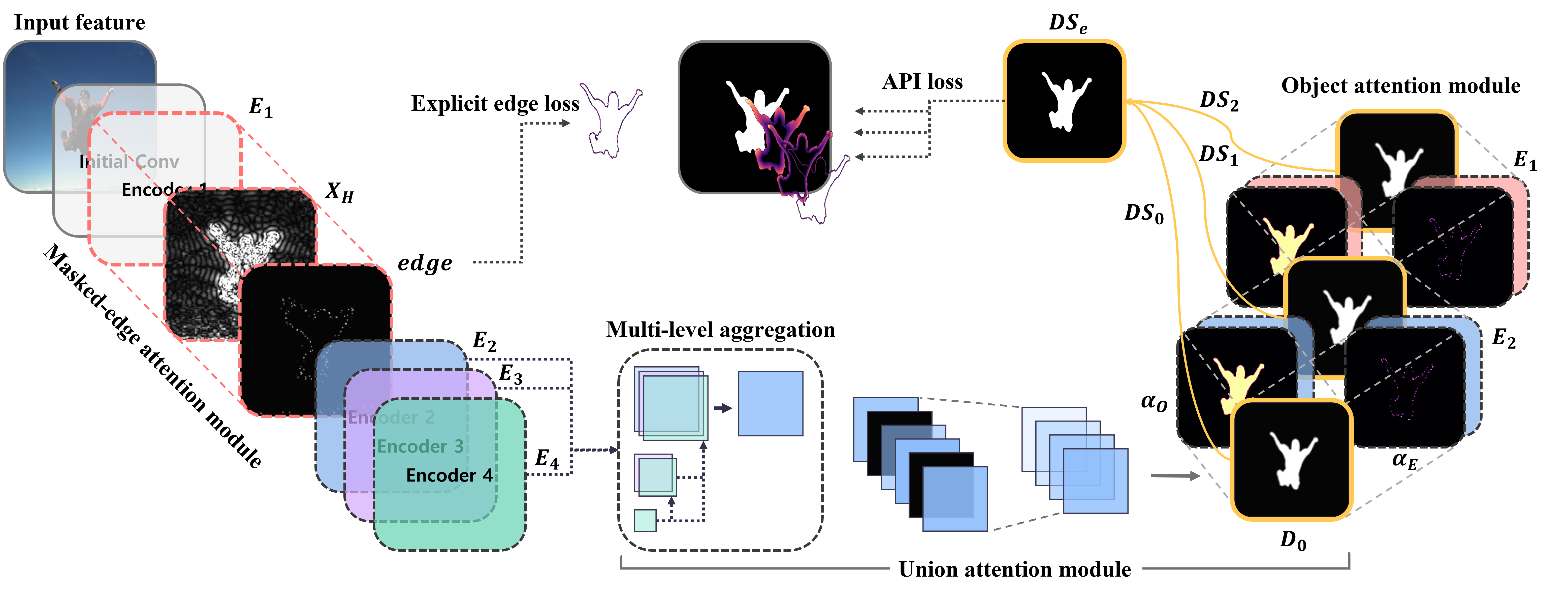}
\end{center}
  \caption{Overview of TRACER architecture.}
\label{fig:long}
\label{fig:onecol}
\label{arch}
\end{figure*}

\subsection{Attention module efficiency}
In CNNs, attention mechanisms have been employed for various tasks to extract important features and enhance the original representation. Images are represented both channel-wise and spatial-wise in CNNs; thus, existing studies have focused on these properties to detect informative features \cite{hu2018squeeze, woo2018cbam}. For the SOD task, PFANet \cite{zhao2019pyramid} employed channel and spatial attention modules separately in shallow and deeper representations, respectively. Another proposed method, PAGENet \cite{wang2019salient}, organized the pyramid attention module that enhances multi-scale representations. In terms of spatial attention, other studies have applied spatial attention weights to each encoder representation to emphasize salient objects \cite{chen2018reverse, liu2018picanet, zhang2018progressive}. However, attention modules are overused in the decoder because existing methods commonly use all the encoder outputs to discriminate salient objects. As a result, computational and memory efficiency is not preserved.


\section{TRACER}
In this section, we describe TRACER, which comprises an efficient backbone encoder along with attention guided salient object tracing modules (i.e., masked edge, union, and object attention modules), as depicted in Fig. \ref{arch}.

\subsection{Architecture overview}\label{backbone}
Because existing backbone encoders, VGG16 (14.7M) and ResNet50 (23.5M), have a vulnerability in feature extraction performance and memory efficiency, an alternative backbone is necessary. Therefore, we employ EfficientNet \cite{tan2019efficientnet} as the backbone encoder and incorporate the existing seven blocks into four blocks in which output resolution is shifted except for the initial convolution blocks. Here, we denote each encoder block output as $E_i$ and the masked edge attention module is initially applied to the first encoder block output $E_1$, which has sufficient boundary representations, to leverage the enhanced edge information and improve memory efficiency. At the decoder, we implement the union and object attention modules, which aggregate multi-level features and incorporate the encoder and decoder outputs, respectively. In the union attention module, we integrate three encoder block outputs $E_2$, $E_3$, and $E_4$, which are obtained by multi-kernel based receptive field blocks, at the scale of $E_2$. The multi-level representations have different distributions; thus, we emphasize more distinct channel and spatial information through the union attention module. The object attention module extracts the distinct object with complementary edge information and utilizes this supplementary information to reduce the discrepancy between the shallow encoder and decoder representations \cite{zhang2018exfuse, pang2019towards}. Moreover, the object attention module consist of depthwise convolution blocks \cite{howard2017mobilenets}, minimizing the number of learning parameters to increase computational efficiency. Finally, TRACER generates four deep supervision maps ($DS_i$) \cite{lee2015deeply}, which are the output the union ($DS_0$), object attention modules ($DS_1$ and $DS_2$), and an ensemble of the $DS$ maps ($DS_e$).

\subsection{Attention guided tracing modules}\label{ATM}
\begin{figure}[ht]
\begin{center}
\includegraphics[scale=0.245]{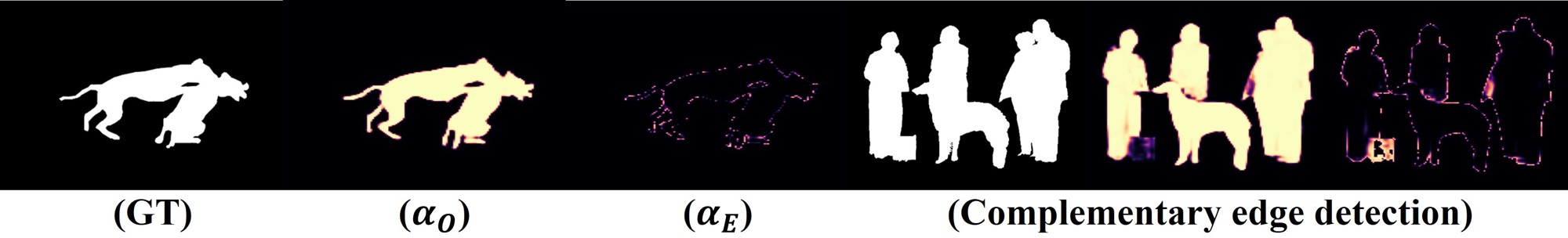}
\end{center}
  \caption{Object and complementary edge detection.}
\label{complementary_edge}
\end{figure}

Detecting distinct objects with edges is critical for improving SOD performance. Using the proposed convolution modules (designed for computational efficiency), we trace the objects and edges through attention guided salient object tracing modules (ATMs) to improve performance.\\
{\bf Masked edge attention: }For tracing edge information, we propose the masked edge attention module (MEAM), which extracts an explicit boundary by employing a fast Fourier transform ($FFT$) \cite{shanmugam1979optimal, xu1996identifying, abdel2003analysis} and enhances the first encoder output boundaries. Existing methods employ the edge information, but they cannot leverage the explicit edges in the feature extraction phases because these methods require the outputs of deeper encoders to obtain the distinct edges. Therefore, we employ the $FFT$ to extract the explicit edge from first encoder representation only. Using $FFT$ and $FFT^{-1}$, the first encoder representation is separated into high and low frequencies as follows:
\begin{equation}
\begin{gathered}
X_{H} = FFT^{-1}(f_{r}^{H}(FFT(X)))
\end{gathered}
\end{equation}
Here, $X$ indicates the input feature, and $FFT(\cdot)$, $FFT^{-1}(\cdot)$ denote the fast Fourier transform and its inverse transform, respectively. Moreover, $f_{r}^{H}(\cdot)$ is a high-pass filter, which eliminates all frequencies except those in radius $r$. To discriminate the explicit edges, we utilize the high frequencies obtained by the high-pass filter, which have sufficient boundary information \cite{haddad1991class, xu1996identifying, wang2020high}. Moreover, $X_{H}$ contains the background noise when $X_{H}$ is transformed from the frequency domain to the spatial domain. Thus, we eliminate noise by applying the receptive field operation $\mathcal{RFB}(\cdot)$ and generate explicit edge as follows: $E=\mathcal{RFB}(X_{H})$. Finally, we compute the edge-refined representation $X_E$ as follows: $X_E = X + E$. Using the $E$, we compute the explicit edge loss.\\
{\bf Union attention: }A union attention module (UAM) is designed to aggregate multi-level features and detect the more important context from both channel and spatial representations. Here, $f(\cdot)$ and $cat(\cdot)$ denote the convolutional operation and channel-wise feature concatenation, respectively. Each encoder output $E_{i \in \{2,3,4\}}$, aggregated to 32, 64, and 128 channels respectively, is integrated as follows:
\begin{equation}
\begin{gathered}
E^{'}_2 = E_2 \otimes f(Up(E_3)) \otimes f(Up(Up(E_4))),\\
E^{''}_3 = f(cat[E_3 \otimes f(Up(E_4)), \;f(Up(E_4))]),\\
E_2^{''}=f(Up(E^{''}_3))
\end{gathered}
\end{equation}
We obtain an aggregated representation, which is the scale of $E_2$, through $X=f(cat[E^{'}_2, \;E^{''}_2]) \in \mathbb{R}^{(32+64+128) \times H_2 \times W_2}$. Following aggregation, it remains that which contextual information is relatively significant in both channel and spatial features. However, existing studies \cite{wang2019salient, zhao2019pyramid} have applied channel and spatial attention modules independently to the decoder and receptive field blocks despite of dependency of both spaces. Thus, we first discriminate the relatively significant channel-wise context and emphasize the spatial information based on complementary confidence scores obtained from the channel context.
\begin{equation}
\begin{gathered}
\label{ca}
\alpha_c = \sigma\left(
\frac{\exp({\mathcal{F}_{q}(\widetilde{X}) (\mathcal{F}_{k}(\widetilde{X}))^{\top}})}
{\sum{\exp(\mathcal{F}_{q}(\widetilde{X}) (\mathcal{F}_{k}(\widetilde{X}))^{\top})}}\mathcal{F}_{v}(\widetilde{X})
\right)
\end{gathered}
\end{equation}
In Eq. \ref{ca}, $\widetilde{X}\in\mathbb{R}^{C\times 1 \times 1}$ is the channel-wise pooled representation, and $\mathcal{F}(\cdot)$ denotes the convolution operation using 1$\times$1 kernel size. Context information is obtained by using the self-attention method and the softmax function to discriminate significant channels $\alpha_c \in \mathbb{R}^{C\times 1 \times 1}$ with a sigmoid function. To refine the aggregated representation $X$, we apply confidence channel weight as follows: $X_{c}=(X \otimes \alpha_{c}) + X$. Subsequently, we retain confidence channels based on the distribution of $\alpha_c$ and the confidence ratio $\gamma$, as follows:
\begin{equation}
\begin{gathered}
\label{channel_mask}
\resizebox{0.8\hsize}{!}{$
\widetilde{X}_{c}=X_c\otimes mask \;\;
\begin{cases}
mask=1, & \mbox{if\;\;} \alpha_c > F^{-1}(\gamma) \\
mask=0, & \mbox{otherwise}
\end{cases}
$}
\end{gathered}
\end{equation}
Here, $F^{-1}(\gamma)$ denotes $\gamma$ quantile of $\alpha_c$. We exclude an area of $\gamma$ in the lower tail of the distribution $\alpha_c$. Then, the refined input $\widetilde{X}_{c}$ is computed spatially to discriminate the salient object and generate the first decoder representation $D_0 \in \mathbb{R}^{1 \times H_2 \times W_2}$, as shown in Eq. \ref{sa}.
\begin{equation}
\begin{gathered}
\label{sa}
D_{0} = 
\frac{\exp({\mathcal{G}_{q}(\widetilde{X}_c) (\mathcal{G}_{k}(\widetilde{X}_c))^{\top}})}
{\sum{\exp(\mathcal{G}_{q}(\widetilde{X}_c) (\mathcal{G}_{k}(\widetilde{X}_c))^{\top})}}\mathcal{G}_{v}(\widetilde{X}_c) 
+ \mathcal{G}_{v}(\widetilde{X}_c)
\end{gathered}
\end{equation}
Here, $\mathcal{G}(\cdot)$ projects the input features to $\widetilde{X}_{c} \in \mathbb{R}^{1 \times H_2 \times W_2}$ using convolutional operation with $1 \times 1$ kernel size. The $D_{0}$ is upsampled to $DS_0$ to obtain a deep supervision map.\\
{\bf Object attention: }To reduce the distribution discrepancy between encoder and decoder representations using minimal parameters, we organize an object attention module (OAM) as a decoder. In contrast to the existing studies \cite{chen2018reverse, zhao2019egnet}, we maintain $D$ as a single channel for decoder efficiency, and the OAM traces both object and complementary edges from each decoder representation $D_i \in\mathbb{R}^{1 \times H \times W}$. To refine the salient object, the object weight $\alpha_{O}$ is calculated as follows: $\alpha_{O}=\sigma(D_{i})$. However, $\alpha_{O}$ cannot always detect the entire object with explicit edge regions; thus, we generate a complementary edge weight $\alpha_{E}$ to cover the undetected regions, as depicted in Fig. \ref{complementary_edge}. For each pixel $x_{ij}$ in $D$, we reverse the detected areas and eliminate background noise corresponding to the denoising ratio $d$ for missed region detection, as shown in Eq \ref{alpha_E}.
\begin{equation}
\begin{gathered}
\label{alpha_E}
\resizebox{0.8\hsize}{!}{$
\alpha_{E}= \;\;
\begin{cases}
0, & \mbox{if\;\;} (-\sigma(x_{ij}) + 1) > d \\
-\sigma(x_{ij}) + 1, & \mbox{otherwise}
\end{cases}
$}
\end{gathered}
\end{equation}
We incorporate the encoder output $E_{i \in \{2, 1 \}}$ and decoder feature $D_{i \in \{0, 1 \}}$, as shown in Eq. \ref{OAM_final_eq}. To reduce discrepancy, we exploit a receptive field operation $\mathcal{RFB}(\cdot)$ and upsample $D_{i+1}$ to generate $DS_{i+1}$.
\begin{equation}
\begin{gathered}
\label{OAM_final_eq}
D_{i+1}=\mathcal{RFB}((\alpha_{O} \otimes E_{2-i}) + (\alpha_{E} \otimes E_{2-i}))
\end{gathered}
\end{equation}

\subsection{Adaptive pixel intensity loss}
\begin{figure}[ht]
\begin{center}
\includegraphics[scale=0.245]{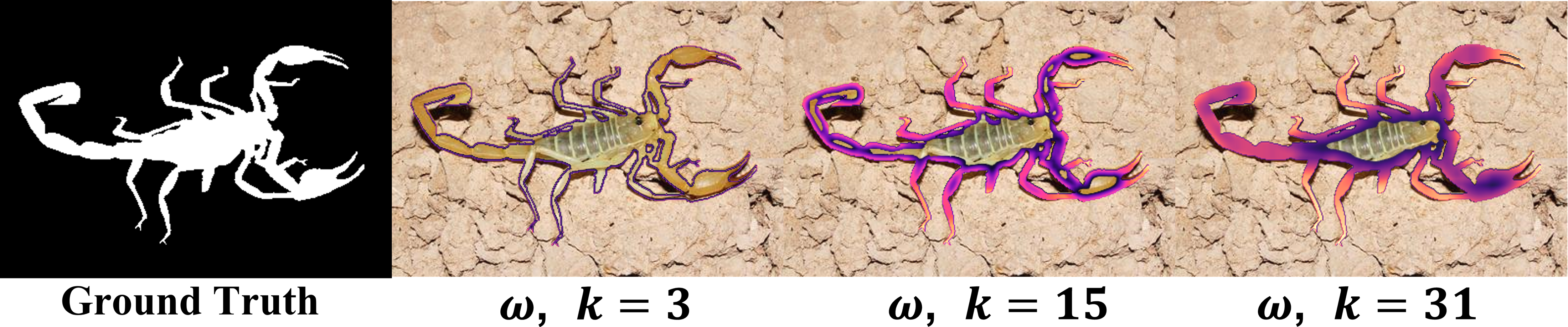}
\end{center}
  \caption{Pixel intensity $\omega$ visualization corresponding to the kernel size $K$.}
\label{loss_viz}
\end{figure}

For a loss function, we combine the binary cross entropy (BCE), IoU, and L1 loss functions to reduce the discrepancy between the object and background. Although the BCE and IoU are globally employed for loss functions, these functions cause a class discrepancy between the foreground and background when all pixels are considered equally. Pixels that are adjacent to fine or explicit edges require more attention compared to pixels in the background and center of the salient object. Thus, we propose adaptive pixel intensity (API) loss, which applies the pixel intensity $\omega$ to each pixel as follows:
\begin{equation}
\begin{gathered}
\label{loss_weight}
\omega_{ij} = (1-\lambda)\sum_{k \in K}{
\left|
\frac{\sum\limits_{h,w \in A_{ij}}{y_{hw}^{k}}}
{\sum\limits_{h,w \in A_{ij}}{1}}
-y_{ij}
\right| y_{ij}}
\end{gathered}
\end{equation}
Here, we aggregate adjacent pixels $(h,w)$ around the target pixel $A_{ij}$ by using multiple kernel size $K$ and excluding weights outside the edges. As depicted in Fig. \ref{loss_viz}, if the target pixel consists of fine edges, multi-kernel aggregation is employed to allocate more intensity to the target pixel than to other pixels. $\lambda$ is an overriding weight that penalizes when employing multi-kernel aggregation because hierarchical aggregation imposes more weights on the pixels at the explicit edges. We empirically set the penalty term $\lambda$ to 0.5, and the kernel size of $K \in \{3,15,31\}$.

The pixel intensity $\omega$ is used for an adaptive BCE (aBCE) loss, as shown in Eq. \ref{abce}. Here, $y$ and $\hat{y}$ denote the label and predicted probability corresponding to binary class $c$, respectively. By using $\omega$, the aBCE loss enables the network to focus more on the local structures related to explicit or fine edges unlike the BCE loss.
\begin{equation}
\begin{gathered}
\label{abce}
\resizebox{1.0\hsize}{!}{$
\mathcal{L}_{BCE}^{a} = 
-\frac{\sum\limits_{i}^{H}{\sum\limits_{j}^{W}}{(1 + \omega_{ij})} \sum\limits_{c=0}^{1}{(y_{c}log(\hat{y}_c)+(1-y_c)log(1-\hat{y}_c))}}
{\sum\limits_{i}^{H}{\sum\limits_{j}^{W}}{(1.5 + \omega_{ij})}}
$}
\end{gathered}
\end{equation}
In contrast, the adaptive IoU loss optimizes the global structures based on the intensive features corresponding to $\omega$. As shown in Eq. \ref{aIoU}, the pixels highly associated with the intensive regions are discriminated and emphasized compared to the original IoU loss.
\begin{equation}
\begin{gathered}
\label{aIoU}
\resizebox{0.8\hsize}{!}{$
\mathcal{L}_{IoU}^{a} = 
1-\frac{\sum\limits_{i}^{H}{\sum\limits_{j}^{W}}{(y_{ij}\hat{y}_{ij})(1+\omega_{ij})}}
{\sum\limits_{i}^{H}{\sum\limits_{j}^{W}}{(y_{ij}+\hat{y}_{ij}-y_{ij}\hat{y}_{ij})(1+\omega_{ij})}}
$}
\end{gathered}
\end{equation}
Moreover, to further improve the network for equivariance learning and to reduce the divergence discrepancy, we measure the L1 distance, which enables the network to learn robustly against noisy labels \cite{ghosh2017robust, wang2020noise}. The L1 loss also deals equally with all pixels; thus, we apply the pixel intensity $\omega$ to the L1 loss to discriminate relatively significant pixels and exclude the noisy pixels adjacent to explicit or fine edges as follows:
\begin{equation}
\begin{gathered}
\label{aL1}
\mathcal{L}_{L1}^{a} = 
\frac{\sum\limits_{i}^{H}{\sum\limits_{j}^{W}}{|y_{ij}-\hat{y}_{ij}|}(1+\omega_{ij})}
{H \times W\sum\limits_{i}^{H}{\sum\limits_{j}^{W}}\omega_{ij}}
\end{gathered}
\end{equation}
To incorporate the above local and global structural intensities, we combine the API loss functions as follows:
\begin{equation}
\begin{gathered}
\label{final_loss}
\resizebox{0.87\hsize}{!}{$
\mathcal{L}_{API}(y, \hat{y}) = \mathcal{L}^{a}_{BCE}(y, \hat{y}) + \mathcal{L}^{a}_{IoU}(y, \hat{y}) + \mathcal{L}^{a}_{L1}(y, \hat{y})
$}
\end{gathered}
\end{equation}
Based on the combined loss function $\mathcal{L}^{API}$, we optimize the final loss using the ground truth $G$, three deep supervisions $DS_{i \in \{0,1,2\}}$, an ensemble of three supervisions $DS_e$, and an explicit edge $E$ obtained from the MEAM, as follows:
\begin{equation}
\begin{gathered}
\mathcal{L} = \sum\limits_{i}\mathcal{L}_{API}(G, DS_i) + \mathcal{L}_{API}(E, \hat{E})
\end{gathered}
\end{equation}


\begin{table*}[t]
\centering
\resizebox{\textwidth}{!}{%
{\Huge
\begin{tabular}{c|c|c|cccc|cccc|cccc|cccc|cccc}
\hline
\multirow{2}{*}{Model} & \multirow{2}{*}{\#Params} & \multirow{2}{*}{GFLOPs} & \multicolumn{4}{c|}{DUTS-TE} & \multicolumn{4}{c|}{DUT-O} & \multicolumn{4}{c|}{HKU-IS} & \multicolumn{4}{c|}{ECSSD} & \multicolumn{4}{c}{PASCAL-S} \\
 &  & & $F_m$ & MAE & $F_A$ & $S_m$ & $F_m$ & MAE & $F_A$ & $S_m$ & $F_m$ & MAE & $F_A$ & $S_m$ & $F_m$ & MAE & $F_A$ & $S_m$ & $F_m$ & MAE & $F_A$ & $S_m$ \\ \hline
BASNet & 87.06M & 195.02 & .860 & .047 & .823 & .866 & .815 & .056 & .767 & .836 & .929 & .032 & .902 & .909 & .944 & .037 & .917 & .916 & .862 & .077 & .821 & .836 \\
CPD-R & 47.85M & 35.54 &.871 & .043 & .821 & .869 & .817 & .056 & .742 & .825 & .927 & .034 & .893 & .906 & .945 & .037 &.913 & .918 & .878 & .072 & .824 & .847 \\
AFNet & 34.3M & - & .873 & .046 & .811 & .866 & .818 & .057 & .734 & .825 & .929 & .036 & .888 & .905 & .941 & .042 & .905 & .913 & .884 & .071 & .830 & .849 \\
EGNet & 112M & 240.3 & .902 & .039 & .836 & .885 & .842 & .053 & .751 & .838 & .943 & .031 & .902 & .918 & .957 & .037 & .919 & .924 & .889 & .075 & .828 & .853 \\
PoolNet & 68.21M & 178 & .901 & .037 & .838 & .886 & .831 & .054 & .742 & .829 & \bf .944 & .030 & .902 & .919 & .956 & .035 & .918 & .926 & .900 & .065 & .842 & .866 \\
SCRN & 25.25M & 30.2 & \bf .906 & .040 & .834 & .885 & \bf .846 & .056 & .750 & .837 & .943 & .033 & .895 & .916 & \bf .961 & .037 & .916 & \bf .927 & \bf .902 & .064 & .838 & \bf .868 \\
MLMSNet & 72.24M & - &.863 & .049 & .791 & .862 & .791 & .063 & .708 & .809 & .927 & .038 & .879 & .907 & .937 & .045 & .890 & .911 & .873 & .074 & .814 & .845 \\
PAGENet & 47.40M & 204 & .834 & .052 & .794 & .855 & .789 & .062 & .743 & .825 & .917 & .037 & .885 & .903 & .932 & .042 & .905 & .912 & .884 & .071 & .830 & .849 \\
MINet & 162M & 174.22 & .894 & .038 & .841 & .884 & .829 & .056 & .752 & .833 & .942 & .029 & .906 & .919 & .954 & .034 & .919 & .925 & .884 & .065 & .833 & .856 \\
ITSD & 26.47M & 31.92 & .893 & .041 & .839 & .885 & .842 & .061 & .767 & \bf .840 & .940 & .031 & .903 & .917 & .953 & .035 & .921 & .925 & .890 & .065 & .839 & .862 \\
GCPANet & 67.06M & 87.98 & .897 & .038 & .836 & .890 & .831 & .057 & .749 & .838 & .942 & .032 & .898 & \bf .920 & .954 & .036 & .912 & \bf .927 & .890 & .063 & .832 & .866 \\
F3Net & 25.54M & 32.86 & .904 & .035 & .851 & .888 & .838 & .053 & .764 & .838 & .943 & \bf .028 & .909 & .917 & .956 & \bf .033 & .925 & .924 & .894 & .062 & .840 & .860 \\
LDF & 25.15M & 31.02 & .903 & \bf .034 & \bf .860 & \bf .892 & .832 & \bf .052 & \bf .768 & .839 & .943 & \bf .028 & \bf .913 & .919 & .955 & .034 & \bf .927 & .925 & .892 & \bf .061 & \bf .846 & .862 \\ \hline

TE0 & 7.45M & 4.26 & .890 & .034 & .849 & .880 & .834 & .049 & .775 & .839 & .935 & .028 & .910 & .913 & .949 & .031 & .923 & .921 & .889 & .059 & .847 & .860\\

TE1 & 9.96M & 4.28 & .893 & .033 & .856 & .885 & .836 & .048 & .781 & .843 & .937 & .027 & .912 & .916 & .951 & .031 & .928 & .923 & .893 & .056 & .855 & .865\\

TE2 & 11.09M & 5.20 & .900 & .030 & .866 & .891 & .841 & .047 & .786 & .846 & .939 & .027 & .915 & .918 & .950 & .031 & .927 & .924 & .894 & .055 & .855 & .866\\

TE3 & 14.02M & 6.26 & .909 & .028 & .873 & .898 & .840 & .046 & .785 & .847 & .944 & .024 & .919 & .922 & .954 & .028 & .930 & .929 & .900 & .053 & .960 & .870\\

TE4 & 20.71M & 8.64 &.915 & .028 & .879 & .902 & .851 & .047 & .798 & .854 & .948 & .023 & .925 & .925 & .958 & .027 & .936 & .931 & .900 & .052 & .862 & .873\\

TE5 & 31.3M & 11.52 & .918 & .026 &	.883 &	.909 &	\bf .853 &	.047 &	\bf .805 & \bf .861 &	.952 & \bf .020 &	.923 &	\bf .933 &	.960 &	\bf .024 &	.938 &	\bf .937 &	.901 &	.048 &	.865 &	.880\\

TE6 & 43.47M & 14.84 & .928 & .025	& .896 &	.915 &	.848 &	.049 &	.797 &	.854 & \bf .954 &	.021 &	.932 &	.932 &	\bf .963 & .025 & \bf .941 & \bf .937 & .908 & .049 & .870 & .881\\

TE7 & 66.27M & 18.86 & \bf .932 & \bf .022 & \bf .904 & \bf .919 & .849 & \bf .045 & .798 & .855 & \bf .954 & \bf .020 & \bf .934 & .932 & .961 & .026 & .940 & .935 & \bf .909 & \bf .047 & \bf .874 & \bf .882\\

\hline
\end{tabular}}}
\caption{Comparison of performance with 13 existing methods on five benchmark datasets. Larger is better in the MaxF ($F_m$) , MeanF ($F_A$), and S-measure metrics, whereas smaller is better in MAE. The best and existing best results are highlighted.}
\label{sota}
\end{table*}

\section{Experiment}
\subsection{Dataset}
We performed the evaluation on five benchmark datasets: DUTS, DUT-OMRON, ECSSD, HKU-IS, and PASCAL-S. DUTS \cite{wang2017learning} is the largest benchmark dataset for SOD. It contains 10,553 training and 5,019 test images. DUT-OMRON \cite{yang2013saliency} has 5,168 images, that include one or more salient objects with relatively complex backgrounds. ECSSD \cite{yan2013hierarchical} comprises 1,000 structurally complex and semantically meaningful scenes. HKU-IS \cite{li2015visual} consist of 4,447 images, that include two or more objects with various backgrounds. PASCAL-S \cite{li2014secrets} has 850 challenging images containing objects of different levels of importance.

\subsection{Experimental setup}
{\bf Hyper-parameters: }We set the batch size to 32 and number of epochs to 100. We used the Adam optimizer with a learning rate of $5\times10^{-5}$ and a weight decay of $10^{-4}$. We observed the validation loss and reduced the learning rate by $10^{-1}$ if the loss did not decrease after five epochs. Additionally, an early stopping operation was applied when the validation loss did not decrease after 10 epochs. We empirically obtained the frequency radius $r=16$ and denoising ratio $d=0.93$ for the masked edge and object attention modules, respectively. To separate insignificant channels in the union attention module, the confident ratio $\gamma$ was set to 0.1. For a fair comparison, we fixed all random seeds to 42.\\
{\bf Implementation details: }We trained TRACER using the DUTS-TR dataset and used the other datasets for testing, following existing studies \cite{wei2020f3net, wei2020label}. In the training phase, data augmentation techniques (e.g., flip, random contrast, and blur) were applied to enhance the network generalization ability. The backbone network weights were initialized with a pre-trained ImageNet dataset. To measure TRACER performance, we used four evaluation metrics MaxF, MeanF, MAE, and S-measure, which are widely adopted \cite{chen2020global, pang2020multi, wei2020f3net, wei2020label}. For the F-measure, $F_\beta$ was calculated from precision-recall pairs, where ${\beta}^2$ was set to 0.3. The S-measure, calculating the object-aware ($S_o$) and region-aware ($S_r$) structural similarities, was calculated as $S_m = \alpha \times S_o + (1-\alpha) \times S_r$, where $\alpha$ = 0.5 \cite{fan2017structure}. TRACER was implemented using the PyTorch framework.

\begin{figure*}[ht]
\begin{center}
\includegraphics[scale=0.52]{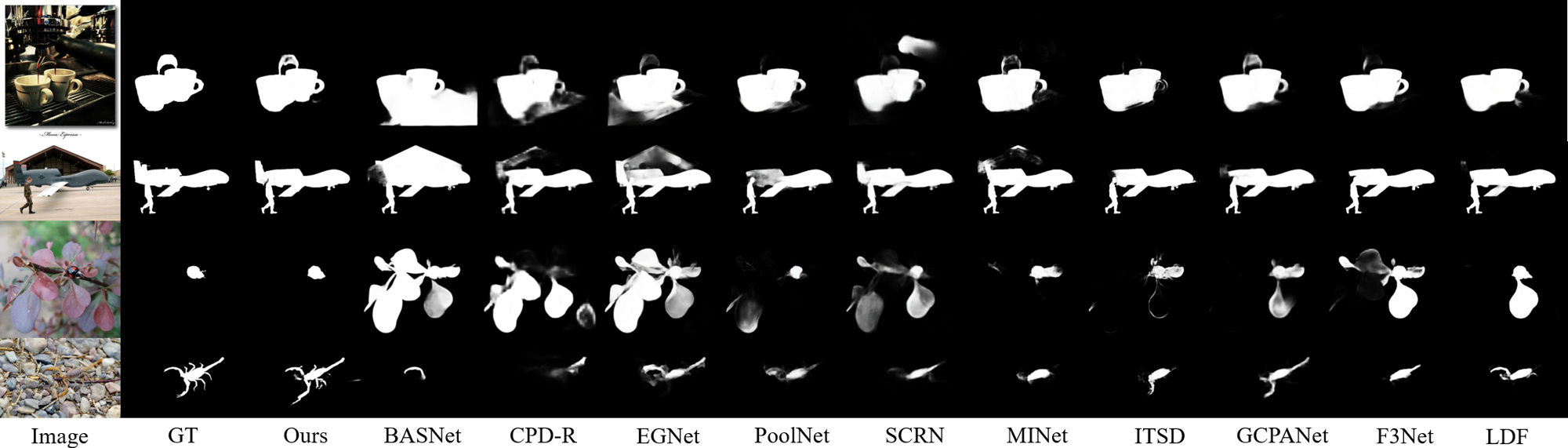}
\end{center}
  \caption{Saliency map comparison of 10 state-of-the-art methods.}
\label{fig:long}
\label{fig:onecol}
\label{visual_comparison}
\end{figure*}

\begin{table*}[t]
\centering
\resizebox{1.0\textwidth}{!}{%
{\huge
\begin{tabular}{c|c|c|c|c|ccc|ccc|ccc|ccc}
\hline
\multirow{2}{*}{Models} & \multirow{2}{*}{Size} & \multirow{2}{*}{\#Params} & \multirow{2}{*}{GFLOPs} & \multirow{2}{*}{MPE} & \multicolumn{3}{c|}{DUTS-TE} & \multicolumn{3}{c|}{DUT-O} & \multicolumn{3}{c|}{HKU-IS} & \multicolumn{3}{c}{ECSSD}\\
& & & & & MAE & $S_m$ & FPS & MAE & $S_m$ & FPS & MAE & $S_m$ & FPS & MAE & $S_m$ & FPS\\ \hline
SCRN & 352 & 25.25M & 30.18 & 10.04m & .040 & .885 & 41.29 & .056 & .837 & 41.52 & .033 & .916 & 44.03 & .037 & .927 & 47.81\\
F3Net & 352 & 25.54M & 32.86 & 7.24m & .035 & .888 & 60.51 & .053 & .838 & 63.22 & .028 & .917 & 78.63 & .033 & .924 & 76.37\\
LDF & 352& 25.15M & 31.02 & 7.05m & .034 & \bf .892 & 64.41 & .052 & .839 & 67.00 & .028 & \bf .919 & 82.89 & .034 & \bf .925 & 79.92\\ \hline

TR-R & 352 & 25.28M & 25.94 & 3.73m & .035 & .890 & 145.48 & .050 &	.845 &	154.38 & .028 & \bf .919 & 154.08 & .033 & \bf .925 & 140.21\\

TE2 & 352 & 11.09M & 5.20 & \bf 2.46m & \bf .030 & .891 & \bf 242.92 & \bf .047 & \bf .846 & \bf 267.25 & \bf .027 & .918 & \bf 260.35 & \bf .031 & .924 & \bf 231.31\\ \hline

\end{tabular}}
}
\caption{Comparison of TRACER effectiveness for ResNet based methods.}
\label{ablation_backbone}
\end{table*}

\subsection{Comparison with state-of-the-art methods}
We compared TRACER with 13 existing methods \cite{feng2019attentive, liu2019simple, qin2019basnet, wang2019salient, wu2019mutual, wu2019cascaded, wu2019stacked, zhao2019egnet, chen2020global, wei2020f3net, wei2020label, pang2020multi, zhou2020interactive}. For a fair comparison, the model parameters, GFLOPs, and saliency maps were obtained from the released code or pre-computed by the authors.\\
{\bf Model performance and efficiency: }As demonstrated in Tab. \ref{sota}, TRACER achieved a state-of-the-art performance and computational efficiency in all the evaluation metrics when compared with the previous 13 methods on the five benchmark datasets. TE2 showed a relatively similar performance compared to LDF, which was the previous outstanding method; however, TE2 required 2.3$\times$ fewer learning parameters and was 6$\times$ faster than LDF. In addition, TE4 was 1.2$\times$ smaller model size than LDF, however, TE4 was 3.6$\times$ faster and showed a significant performance improvement than LDF. Based on the model parameters, GCPANet (67.06M) and PoolNet (68.21M) required a similar level of learning parameters compared with TE7 (66.27M); however, TE7 significantly outperformed other methods by a large margin, that is, it was 4.6$\times$ and 9.4$\times$ faster, respectively.\\
{\bf Saliency map comparison: }We evaluated TRACER with 10 previous methods \cite{qin2019basnet, wu2019cascaded, wu2019stacked, zhao2019egnet, chen2020global, liu2019simple, pang2020multi, wei2020f3net, wei2020label, zhou2020interactive} by visualizing the saliency maps. As shown in Fig. \ref{visual_comparison}, we sampled large-scale (rows 1 and 2) and small-scale salient objects with relatively complex backgrounds (rows 3 and 4). All the methods detected the large-scale object; however, the existing methods could not obtain the complete detailed region. For the detection of small-scale objects, TRACER could extract the salient objects precisely; in contrast, the previous methods identified objects that included the background or only a few areas.


\begin{table}[t]
\centering
\resizebox{0.47\textwidth}{!}{%
{\huge
\begin{tabular}{c|cc|cc|cc|cc}
\hline
\multirow{2}{*}{Modules} & \multicolumn{2}{c|}{DUTS-TE} & \multicolumn{2}{c|}{DUT-O} & \multicolumn{2}{c|}{HKU-IS} & \multicolumn{2}{c}{ECSSD}\\ 
& MAE & $S_m$ & MAE & $S_m$ & MAE & $S_m$ & MAE & $S_m$\\ \hline

--$\dagger$ & .038 & .874 & .059 &  .822 & .033 &.905 & .042 & .907\\

MEAM & .035 & .876 & .059 & .831 & .030 & .910 & .042 & .915\\

OAM & .034 & .880 &.057 & .828 & .027 & .912 & .036 & .917\\ 

UAM & .027 & .909 & .056 & .838 & .026 & .929 & .032 & .927\\ \hline

UAM+MEAM & .027 & .911  &.053 & .844 & .024 & .926 & .030 & .933\\
UAM+OAM & .024 & .913 & .048 & .842 & .022 & .928 & .027 & .934\\

UA+MEA+OA & \bf .022 & \bf .919 & \bf .045 & \bf .855 & \bf .020 & \bf .932 & \bf .026 & \bf .935\\\hline
\end{tabular}}}
\caption{Comparison of the TE7 performance for the combination of ATMs. $\dagger$ indicates that all the ATMs are eliminated.}
\label{ablation_module}
\end{table}

\section{Ablation study}
We have demonstrated that TRACER improves SOD performance substantially with respect to memory efficiency by employing a smaller backbone and attention guided tracing modules (ATMs). To evaluate the effectiveness of TRACER framework and subsequently enhance model explainability, we conducted ablation studies on TRACER.\\
{\bf TRACER framework effectiveness: }We organized the backbone encoder with EfficientNet because the existing backbone encoder had unsatisfactory feature extraction performance and memory efficiency. To evaluate the effectiveness of the TRACER framework, we compared it with existing methods \cite{wu2019stacked, wei2020f3net, wei2020label}, which showed outstanding performance. For a fair comparison, we measured model training and inference times, that is, minutes per epoch (MPE) and frames per second (FPS), respectively, under the same conditions. We adopted TRACER-ResNet (TR-R) and TE2, which have the same backbone or input size. As listed in the Tab.\ref{ablation_backbone}, the TE2 performed at least 2.9$\times$ to maximum 4.1$\times$ faster than the existing methods on training each epoch and at least 3.8$\times$ to maximum 6.4$\times$ faster on inference times. Indeed, TR-R occupied approximately 12.9\% of of total GFLOPs at the decoder structures. In contrast, the multi-decoder frameworks proposed from the existing methods occupied 32.6\% (SCRN), 38.2\% (F3Net), and 34.6\% (LDF) of total GFLOPs, respectively. As a result, we observed that when we employed a conventional backbone, the TRACER framework improved network efficiency.\\
{\bf ATMs effectiveness: }To validate the performance improvement originating from proposed attention modules, we eliminated the ATMs in TE7. In Tab. \ref{ablation_module}, we observed a change in performance improvement with and without ATMs. In particular, when we eliminated the UAM, TRACER performance decreased significantly (rows 1 to 4). Because the UAM enhances the aggregated multi-level representation, which contains various levels of importance, the UAM contributed more to the improvement than the MEAM and OAM. It means that discriminating distinct features from the aggregated multi-level representation, neither the edge representation leverage nor decoder structure, determined much of the SOD performance. Based on the UAM application, employing the OAM outperformed the MEAM result (rows 5 to 6). That is, the decoder structure secondly contributed to the performance gain because it complemented fine edges and undetected regions by aggregating encoder representations.\\
{\bf Adaptive pixel intensity: }In Tab. \ref{ablation_api}, the proposed API loss functions (rows 5 to 7) displayed outstanding local-global structure awareness and network robustness against noisy labels, more so than other loss function combinations. In rows 2 and 3, we observed that using API was more effective for local and global structure awareness than the weighted BCE and IoU losses proposed by the existing study \cite{wei2020f3net}. When we additionally adopted L1 loss (row 4), MAE improved while preserving $S_m$ performance. In particular, applying API to MAE yielded a higher performance improvement than the conventional loss combination. Moreover, we examined the performance variation corresponding to the penalty term $\lambda$. The gain was unsatisfactory at a higher $\lambda$ because it highly penalized $\omega$ assigned to the pixels adjacent to the fine edges. When we fully reflected the $\omega$ on the pixels, it showed limited performance gain of the local-global structure awareness and robustness because of weight overwhelming.

\begin{table}[t]
\centering
\resizebox{.47\textwidth}{!}{%
{\huge
\begin{tabular}{c|cc|cc|cc|cc}
\hline
\multirow{2}{*}{Loss functions} & \multicolumn{2}{c|}{DUTS-TE} & \multicolumn{2}{c|}{DUT-O} &
\multicolumn{2}{c|}{HKU-IS} & \multicolumn{2}{c}{ECSSD}\\
& MAE & $S_m$ & MAE & $S_m$ & MAE & $S_m$ & MAE & $S_m$\\
\hline
BCE+IoU                       & .038 & .895 & .055 & .835 & .034 & .915 & .035 & .923\\
wBCE+wIoU                     & .034 & .896 & .053 & .839 & .033 & .920 & .034 & .925\\
$\omega$(BCE+IoU)$^{\dagger}$ & .030 & .909 & .051 & .851 & .031 & .929 & .032 & .930\\
\hline
BCE+IoU+L1                    & .027 & .896 & .049 & .836 & .027 & .917 & .030 & .925\\

API ($\lambda$=0.9)$^{\dagger}$ & .026 & .906 & .048 & .840 & .023 & .920 & .030 & .927\\
API ($\lambda$=0.5)$^{\dagger}$ & \bf .022 & \bf .919 & \bf .045 & \bf .855 & \bf .020 & \bf .932 & \bf .026 & \bf .935\\
API ($\lambda$=0.0)$^{\dagger}$ & .024 & .912 & .047 & .849 & \bf .020 & .930 & .027 & .931\\
\hline
\end{tabular}}}
\caption{Comparison of effects on the adaptive pixel intensity in the loss function. $\dagger$ indicates API loss function.}
\label{ablation_api}
\end{table}



\section{Conclusion}
We studied the inefficiencies in the existing encoder-decoder structure to improve the SOD performance along with network efficiency. We proposed TRACER, which discriminates salient objects by employing ATMs. TRACER detects the objects and edges in both channel and spatial-wise representations using minimal learning parameters. To treat the relative importance of pixels, we propose an adaptive pixel intensity loss function. For model explainability, we observe the model GFLOPs using the effectiveness of ATMs, through ablation studies. TRACER improves the performance and computational efficiency in all evaluation metrics in comparison to the 13 existing methods on the five benchmark datasets.


\section*{Acknowledgements}
This research was supported by National Research Foundation of Korea (NRF-2019R1F1A1060250). This research was also supported by Brain Korea 21 FOUR.


\section*{Supplementary}
\subsection{Convolutional modules}
\begin{figure}[ht]
\begin{center}
\includegraphics[scale=0.47]{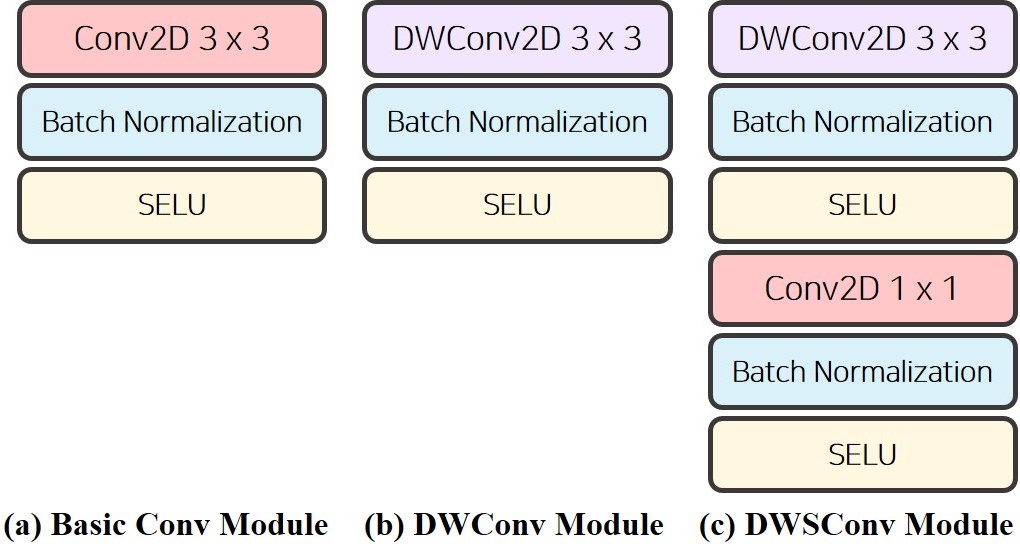}
\end{center}
  \caption{Convolutional modules applied in TRACER.}
\label{fig:long}
\label{fig:onecol}
\label{conv_module}
\end{figure}

\begin{figure}[t]
\begin{center}
\includegraphics[scale=0.57]{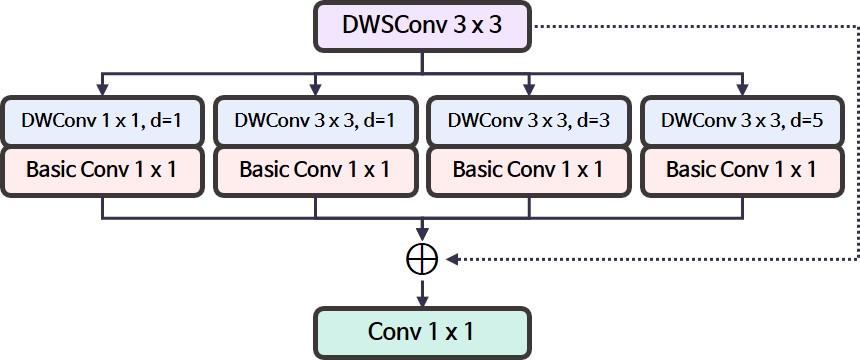}
\end{center}
  \caption{Overview of depthwise dilated receptive field module.}
\label{fig:long}
\label{fig:onecol}
\label{rfb}
\end{figure}

{\bf Basic convolutional module: }The attention guided salient object tracing modules mainly comprise three convolutional modules, as depicted in Fig. \ref{conv_module}. We design the post-activation structure employed in ResNet \cite{he2016deep, he2016identity} to regularize a network and leverage a generalization effect. The original structure used ReLU, whereas our basic convolutional module adopts SELU \cite{klambauer2017self}, which can control negative values and make networks more robust than ReLU. To achieve computational efficiency, as demonstrated by a depthwise convolutional (DWConv) in a MobileNet \cite{howard2017mobilenets}, we also reorganize the DWConv and depthwise separable convolutional (DWSConv) modules by applying post-activation after each convolutional layer.
\\
{\bf Depthwise dilated receptive field module: }Existing receptive field blocks \cite{liu2018receptive} are organized with naive convolutional combinations; however, these combinations are inefficient in larger channels. Thus, we design a depthwise dilated receptive field module (DDRM) based on three convolutional modules, which are BasicConv, DWConv, and DWSConv modules, as illustrated in Fig. \ref{rfb}. Although depthwise convolutional has a lower performance than naive convolutional modules \cite{howard2017mobilenets}, parameter reduction is effective in large channels. We employ the DDRM in the masked edge attention and object attention modules, which require receptive representations and large computational resources (e.g., resolution or number of channels).

\begin{table}[ht]
\centering
\resizebox{.47\textwidth}{!}{%
{\huge
\begin{tabular}{c|cc|cc|cc|cc}
\hline
\multirow{2}{*}{$\gamma$} & \multicolumn{2}{c|}{DUTS-TE} & \multicolumn{2}{c|}{DUT-O} & \multicolumn{2}{c|}{HKU-IS} & \multicolumn{2}{c}{ECSSD}\\
& MAE & $S_m$ & MAE & $S_m$ & MAE & $S_m$ & MAE & $S_m$\\ \hline
-$^*$ & 0.025 & 0.916 & 0.049 & 0.853 & 0.022 & 0.930 & 0.027 & \bf 0.935\\
0.1 & \bf 0.022 & \bf 0.919 & \bf 0.045 & \bf 0.855 & \bf 0.020 & \bf 0.932 & \bf 0.026 & \bf 0.935\\
0.5 & 0.025 & 0.918 & 0.048 & 0.849 & 0.022 & 0.930 & 0.028 & 0.934\\
0.7 & 0.026 & 0.909 & 0.047 & 0.835 & 0.023 & 0.927 & 0.029 & 0.931\\
0.9 & 0.028 & 0.899 & 0.048 & 0.821 & 0.025 & 0.920 & 0.031 & 0.926\\ \hline

\end{tabular}}}
\caption{Comparison of TRACER performance with respect to confidence ratio $\gamma$. * indicates the condition in which confidence ratio was not applied.}
\label{ablation_maksing}
\end{table}

\subsection{Confidence ratio}
To observe the confidence channel effectiveness with a confidence ratio $\gamma$, we calibrated $\gamma$ as depicted in Tab. \ref{ablation_maksing}. The network performance significantly decreased as $\gamma$ increased. Because channels were significantly eliminated when using the large values of $\gamma$ for detecting distinct representations, we obtained an unsatisfactory performance with large values of $\gamma$. Moreover, removal of the complementary masking resulted in poor performance compared to when masking was applied.

\subsection{Edge leverage}
We observed the effect of the location of the MEAM on performance, as listed in Tab. \ref{ablation_MEAM}. When the MEAM was located at the end of the deeper encoder, TRACER performance decreased gradually, and this MEAM in the deeper encoder resulted in memory and computational inefficiency. Because the deeper encoder output contains a more abstract representation than the shallow encoder, the MEAM is ineffective on deeper outputs. Moreover, it cannot propagate an edge-refined representation to the downstream feature extraction. Therefore, positioning the MEAM in the shallow encoder improved performance and network efficiency.

\begin{table}[ht]
\centering
\resizebox{.47\textwidth}{!}{%
{\Huge
\begin{tabular}{c|c|c|cc|cc|cc|cc}
\hline
\multirow{2}{*}{E$_{i}$} & \multirow{2}{*}{\#Params} & \multirow{2}{*}{GFLOPs} & \multicolumn{2}{c|}{DUTS-TE} & \multicolumn{2}{c|}{DUT-O} & \multicolumn{2}{c|}{HKU-IS} & \multicolumn{2}{c}{ECSSD}\\  
& & & MAE & $S_m$ & MAE & $S_m$ & MAE & $S_m$ & MAE & $S_m$\\ \hline
1 & 66.27M & 18.86 & \bf .022 & \bf .919 & \bf .045 & \bf .855 & \bf .020 & \bf .932 & \bf .026 & \bf .935\\
2 & 66.29M & 18.72 & .025 & .916 & .048 & .847 & .023 & .924 & \bf .026 & \bf .935\\
3 & 66.56M & 18.86 & .027 & .913 & .051 & .841 & .024 & .922 & .028 & .931\\
4 & 68.74M & 19.18 & .028 & .912 & .053 & .839 & .024 & .920 & .029 & .929\\

\hline
\end{tabular}}}
\caption{Comparison of the effects of different locations of the MEAM.}
\label{ablation_MEAM}
\end{table}

\subsection{Frequency radius and denoising ratio}
We calibrated the frequency radius $r$ and denoising ratio $d$, which were used in the masked edge attention and object attention modules. As shown in Tab. \ref{ablation_freq_denoise}, we found that when $r=16$ and $d=0.93$, TRACER performed optimally. When $r$ was between 14 and 18, TRACER exhibited robust performance. Because $d$ clarified edges with undetected regions in the OAM, we obtained an unsatisfactory MAE performance as $d$ decreased. Moreover, high $d$ values also excluded the undetected regions, resulting in poor TRACER performance.

\begin{table}[t]
\centering
\resizebox{.47\textwidth}{!}{%
{\huge
\begin{tabular}{c|cc|cc|cc|cc}
\hline
\multirow{2}{*}{$r$} & \multicolumn{2}{c|}{DUTS-TE} & \multicolumn{2}{c|}{DUT-O} & \multicolumn{2}{c|}{HKU-IS} & \multicolumn{2}{c}{ECSSD}\\
& MAE & $S_m$ & MAE & $S_m$ & MAE & $S_m$ & MAE & $S_m$\\ \hline
10 & 0.025 & 0.916 & 0.047 & 0.851 & 0.022 & 0.929 & 0.029 & 0.932\\
12 & 0.023 & 0.917 & 0.047 & 0.853 & 0.021 & 0.931 & 0.027 & 0.934\\
14 & \bf 0.022 & 0.916 & 0.046 & \bf 0.855 & 0.021 & \bf 0.932 & 0.027 & \bf 0.936\\
16 & \bf 0.022 & \bf 0.919 & \bf 0.045 & \bf 0.855 & \bf 0.020 & \bf 0.932 & \bf 0.026 & 0.935\\
18 & 0.023 & 0.918 & \bf 0.045 & 0.853 & 0.021 & \bf 0.932 & \bf 0.026 & 0.935\\
20 & 0.023 & 0.917 & 0.046 & 0.854 & \bf 0.020 & 0.930 & 0.027 & 0.933\\
22 & 0.024 & 0.917 & 0.048 & 0.852 & 0.022 & 0.931 & 0.028 & 0.934\\ \hline

\multirow{2}{*}{$d$} & \multicolumn{2}{c|}{DUTS-TE} & \multicolumn{2}{c|}{DUT-O} & \multicolumn{2}{c|}{HKU-IS} & \multicolumn{2}{c}{ECSSD}\\
& MAE & $S_m$ & MAE & $S_m$ & MAE & $S_m$ & MAE & $S_m$\\ \hline
0.87 & 0.029 & 0.913 & 0.050 & 0.852 & 0.023 & 0.930 & 0.028 & 0.933\\
0.89 & 0.025 & 0.916 & 0.048 & 0.853 & 0.021 & \bf 0.932 & 0.028 & 0.934\\
0.91 & 0.023 & 0.917 & 0.046 & \bf 0.855 & 0.021 & 0.930 & 0.027 & 0.933\\
0.93 & \bf 0.022 & \bf 0.919 & \bf 0.045 & \bf 0.855 & \bf 0.020 & \bf 0.932 & \bf 0.026 & \bf 0.935\\
0.95 & 0.024 & 0.916 & 0.047 & 0.853 & 0.022 & 0.931 & \bf 0.026 & 0.934\\
0.97 & 0.025 & \bf 0.919 & 0.046 & \bf 0.855 & 0.021 & \bf 0.932 & 0.027 & \bf 0.935\\
0.99 & 0.028 & 0.917 & 0.049 & 0.853 & 0.022 & 0.930 & 0.027 & 0.931\\ \hline

\end{tabular}}}
\caption{Comparison of frequency radius $r$ and denoising ratio $d$. The best results in each experiment are highlighted.}
\label{ablation_freq_denoise}
\end{table}

\subsection{Channel and spatial dependency}
This study has addressed that channel and spatial-wise representations could be complementary. To verify the effectiveness of a complementary attention structure, we organized both modules parallel. Tab. \ref{ablation_uam} demonstrated the complementary structure outperformed the parallel structure. Therefore, the complementary structure, which eliminated relatively insignificant channels in the spatial attention module following the channel attention weights, was effective compared to the parallel structure.

\begin{table}[t]
\centering
\resizebox{.47\textwidth}{!}{%
{\Huge
\begin{tabular}{c|cc|cc|cc|cc}
\hline
\multirow{2}{*}{Modules} & \multicolumn{2}{c|}{DUTS-TE} & \multicolumn{2}{c|}{DUT-O} & \multicolumn{2}{c|}{HKU-IS} & \multicolumn{2}{c}{ECSSD}\\
& MAE & $S_m$ & MAE & $S_m$ & MAE & $S_m$ & MAE & $S_m$\\ \hline
UAM-parallel & .027 & .914 & .050 & .851 & .023 & .930 & .030 & .932\\
UAM & \bf .022 & \bf .919 & \bf .045 & \bf .855 & \bf .020 & \bf .932 & \bf .026 & \bf .935\\ \hline
\end{tabular}}}
\caption{Comparison of channel and spatial-wise attention structures in UAM.}
\label{ablation_uam}
\end{table}

\begin{figure}[t]
\begin{center}
\includegraphics[scale=0.125]{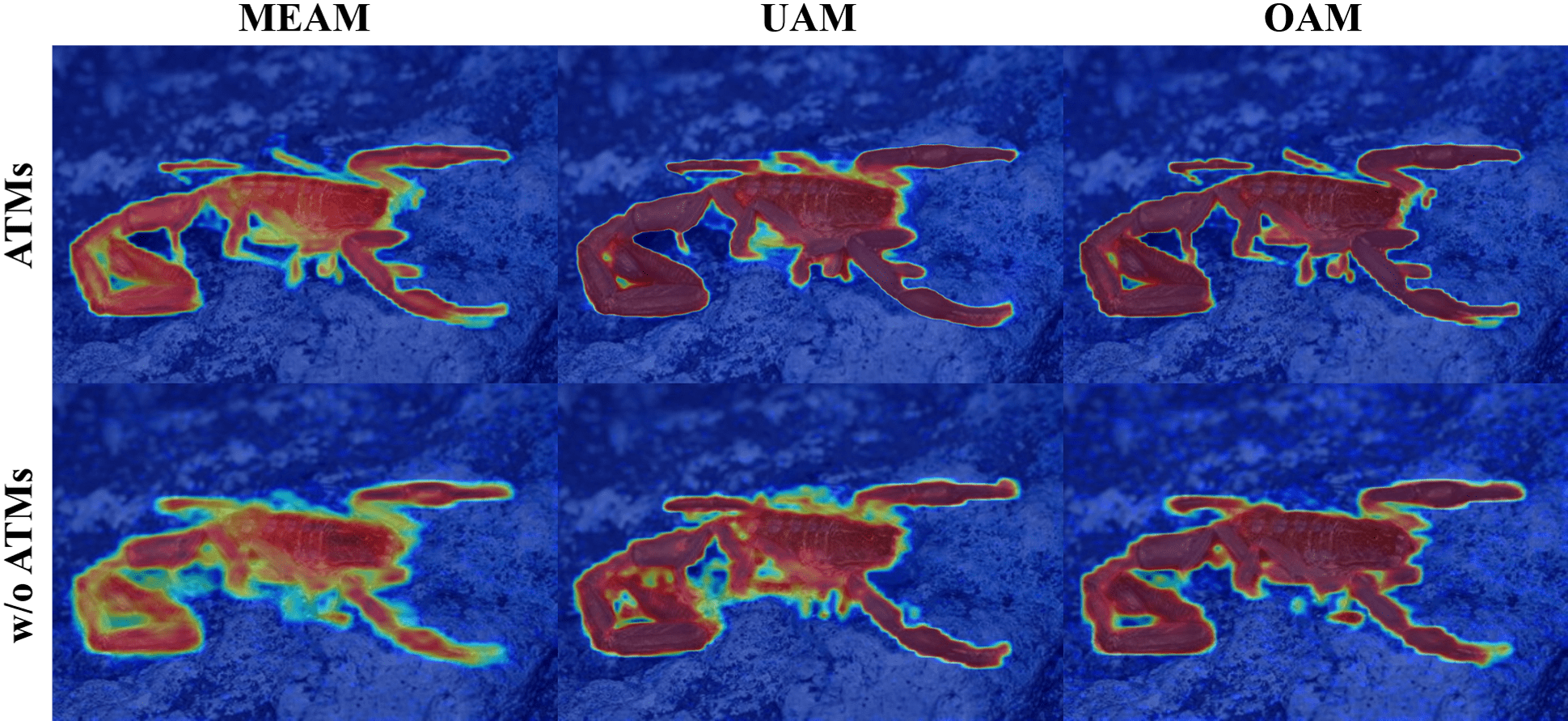}
\end{center}
\caption{Comparison of heatmap visualization corresponding to the application of ATMs. The first row indicates the heatmaps obtained from each module.}
\label{heatmap_viz}
\end{figure}

\subsection{Qualitative ATMs comparison}
We visualized heatmaps and compared the results corresponding to the application of ATMs for the module explainability, as depicted in Fig \ref{heatmap_viz}. When the ATMs are applied, the network not only discriminates the fine edges and redundant regions but also shows robustness against noise.

\subsection{Weighted adaptive pixel intensity loss combination}
We experimented with a weighted combination of the proposed adaptive pixel intensity (API) loss function to observe the influence on local-global structure awareness and robustness against noisy labels. As listed in Tab. \ref{ablation_loss_combination}, when we imposed more weights on adaptive binary cross entropy (aBCE), we obtained F-measure (i.e., MaxF and MeanF) improvement but the MAE and $S_m$ were unsatisfactory. In contrast, although higher weights on adaptive IoU displayed outstanding local-global structure awareness ($S_m$) compared to other combinations, they could not handle noisy labels. Indeed, some combinations displayed outstanding performance at specific measures compared to the equally weighted combination on DUTS-TE and DUT-O. However, the equally weighted combination, (1.0*$\mathcal{L}_{BCE}^{a}$ + 1.0*$\mathcal{L}_{IoU}^{a}$ + 1.0*$\mathcal{L}_{L1}^{a}$), was adopted as it achieved satisfactory generalization performance.

\begin{table}[t]
\centering
\resizebox{0.47\textwidth}{!}{%
{\Huge
\begin{tabular}{c|cccc|cccc}
\hline
\multirow{2}{*}{$\mathcal{L}_{BCE}^{a}$ / $\mathcal{L}_{IoU}^{a}$ / $\mathcal{L}_{L1}^{a}$} & \multicolumn{4}{c|}{DUTS-TE} & \multicolumn{4}{c}{DUT-O}\\
& $F_m$ & MAE & $F_A$ & $S_m$ & $F_m$ & MAE & $F_A$ & $S_m$\\ \hline
1.0 / 0.5 / 0.5 & \bf .935 & .025 & \bf .907 & .914 & .848 & .047 & .796 & .856\\
0.5 / 1.0 / 0.5 & .932 & .027 & .905 & \bf .923 & .846 & .049 & .795 & \bf .859\\
0.5 / 0.5 / 1.0 & .927 & .023 & .901 & .914 & .846 & \bf .045 & .794 & .851\\ \hline
1.0 / 1.0 / 1.0 & .932 & {\bf .022} & .904 & .919 & {\bf .849} & {\bf .045} & \bf .798 & .855\\ \hline

\end{tabular}}}
\caption{Comparison of weighted API loss combinations.}
\label{ablation_loss_combination}
\end{table}

\begin{table*}[t]
\centering
\resizebox{\textwidth}{!}{%
{\Huge
\begin{tabular}{c|c|c|cccc|cccc|cccc|cccc}
\hline
\multirow{2}{*}{Models} & \multirow{2}{*}{\#Params} & \multirow{2}{*}{GFLOPs} & \multicolumn{4}{c|}{DUTS-TE} & \multicolumn{4}{c|}{DUT-O} & \multicolumn{4}{c|}{HKU-IS} & \multicolumn{4}{c}{ECSSD}\\
& & & MaxF & MAE & MeanF & $S_m$ & MaxF & MAE & MeanF & $S_m$ 
& MaxF & MAE & MeanF & $S_m$ & MaxF & MAE & MeanF & $S_m$\\ \hline

TR-ResNet50 & 25.28M & 25.94 & 0.905 & 0.035 & 0.855 & 0.890 & 0.849 & 0.050 & 0.776 & 0.845 & 0.943 & 0.028 & 0.910 & 0.919 & 0.955 & 0.033 & 0.925 & 0.925\\

TR-DenseNet201 & 19.28M & 22.26 & 0.891 & 0.039 & 0.840 & 0.883 & 0.837 & 0.053 & 0.761 & 0.837 & 0.931 & 0.033 & 0.895 & 0.911 & 0.949 & 0.038 & 0.912 & 0.920\\

TR-ResNeXt50 & 24.75M & 26.68 & 0.904 & 0.035 & 0.853 & 0.891 & 0.845 & 0.052 & 0.774 & 0.844 & 0.942 & 0.028 & 0.907 & 0.920 & 0.956 & 0.032 & 0.924 & 0.926\\

TR-Res2Net50 & 27.49M & 27.98 & 0.910 & 0.033 & 0.860 & 0.895 & 0.844 & 0.050 & 0.765 & 0.840 & 0.942 & 0.028 & 0.905 & 0.920 & 0.956 & 0.035 & 0.920 & 0.923\\

TR-Xception & 21.99M & 25.38 & 0.906 & 0.032 & 0.859 & 0.897 & 0.846 & 0.050 & 0.775 & 0.845 & 0.944 & 0.027 & 0.910 & 0.921 & 0.955 & 0.032 & 0.924 & 0.927\\
\hline

TE4 & 20.71M & 8.64 & \bf 0.915 & \bf 0.028 & \bf 0.879 & \bf 0.902 & \bf 0.851 & \bf 0.047 & \bf 0.798 & \bf 0.854 & \bf 0.948 & \bf 0.023 & \bf 0.925 & \bf 0.925 & \bf 0.958 & \bf 0.027 & \bf 0.936 & \bf 0.931\\ \hline

\end{tabular}}}
\caption{Comparison of TRACER framework effectiveness on other backbones.}
\label{other_backbones}
\end{table*}

\begin{table}[ht]
\centering
\resizebox{0.473\textwidth}{!}{%
{\Huge
\begin{tabular}{ccccc|cc|cc|cc|cc}
\hline
\multicolumn{5}{c|}{Kernel size} & \multicolumn{2}{c|}{DUTS-TE} & \multicolumn{2}{c|}{DUT-O} & \multicolumn{2}{c|}{HKU-IS} & \multicolumn{2}{c}{ECSSD}\\
{3 \hspace{0pt plus 2filll}} & {9 \hspace{0pt plus 2filll}} & {15 \hspace{0pt plus 2filll}} & {23 \hspace{0pt plus 2filll}} & {31 \hspace{0pt plus 2filll}} & MAE & $S_m$ & MAE & $S_m$ & MAE & $S_m$ & MAE & $S_m$\\ \hline

\checkmark & & & &            & .024 & .905 & .048 & .842 & .023 & .923 & .030 & .927\\
& & \checkmark & &            & .023 & .910 & .047 & .845 & .022 & .925 & .029 & .927\\
& & & &  \checkmark           & .023 & .916 & .046 & .851 & .021 & .930 & .027 & .929\\ 

\checkmark & & \checkmark & & & .024 & .909 & .046 & .849 & .022 & .926 & .029 & .930\\
& & \checkmark & & \checkmark & .023 & .915 & .046 & .852 & .021 & .931 & \bf .026 & .934\\

\checkmark & & \checkmark & & \checkmark & \bf .022 & .919 & .045 & \bf .855 & \bf .020 & .932 & \bf .026 & .935\\
\checkmark & \checkmark & \checkmark & \checkmark & \checkmark & \bf .022 & \bf .921 & \bf .044 & .854 & .021 & \bf .933 & \bf .026 & \bf .936\\ 
\hline
\hline

\end{tabular}}}
\caption{Comparison of TRACER performance with respect to multiple kernel sizes in API loss.}
\label{ablation_kernel_size}
\end{table}

\subsection{Hierarchical multi-kernel aggregation effect}
We examined the effectiveness of using multiple kernel sizes $K$ for API loss. As listed in Tab. \ref{ablation_kernel_size}, employing hierarchical multi-kernel aggregation (rows 6 and 7) displayed outstanding performance compared to single or bi-kernel aggregation. Indeed, as shown in rows 1 to 3, increasing the kernel size enabled the network to learn local and global structure awareness. Based on the kernel size increase, we observed that the capacity of noisy label robustness and structural awareness improved when we incorporated multiple kernel sizes. Although adopting multi-kernel aggregation was more effective than single or bi-kernel aggregation, performance improvement was limited when we applied five different kernel sizes. Thus, hierarchical aggregation with three kernels (e.g., $K \in \{3,15,31\}$) could be empirically obtained for a reasonable level of API loss.

\subsection{Framework Effectiveness}
To observe the TRACER framework effectiveness, we experimented with other alternative backbones that had similar model sizes. Following the encoder aggregation strategy as applied in TRACER, we integrated the backbones in four encoder blocks. Since TRACER required a different input size for each encoder output ($E_2$ to $E_4$), we modified the last dense block in DenseNet by employing a pooling operation. As depicted in Tab. \ref{other_backbones}, DenseNet \cite{huang2017densely} required fewer parameters and demonstrated suitable computational efficiency, although the modifications made in the dense block4 for applying TRACER resulted in an unsatisfactory performance. In contrast, Xception \cite{chollet2017xception} achieved outstanding performance with remarkable network efficiency when compared to that of existing methods and other alternative backbones; however, this Xception backbone was not able to outperform TE4, which had significant performance improvements and was 3$\times$ faster.

\subsection{Attention guided object tracing}
\begin{figure}[ht]
\begin{center}
\includegraphics[scale=0.31]{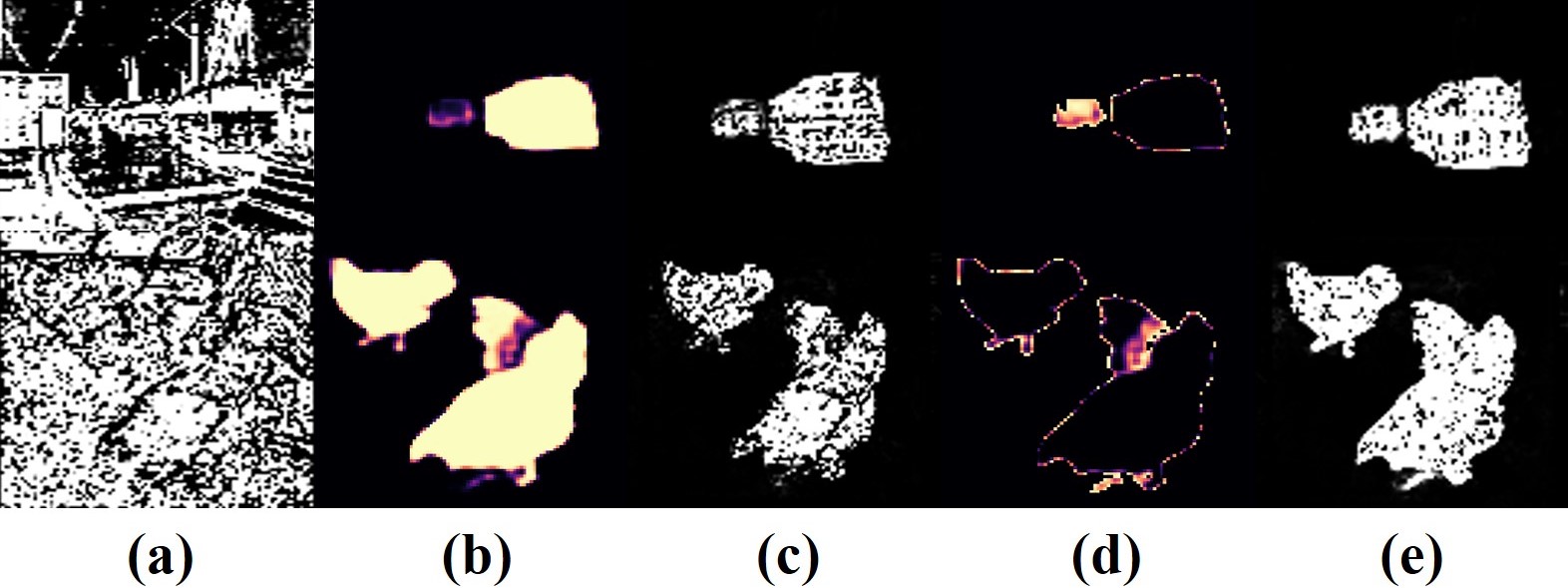}
\end{center}
  \caption{Input feature and attention map visualization. The feature maps are sampled from the original channels.}
\label{fig:long}
\label{fig:onecol}
\label{ablation_OAM}
\end{figure}

As the OAM enhanced salient object with undetected regions, we visualized object and complementary edge attention maps to observe the OAM effectiveness as depicted in Fig. \ref{ablation_OAM}. The input feature $X$-(a), which was obtained from the encoder outputs, was clarified by the object attention $\alpha_O$-(b) and transformed to (c). Then, we obtained the complementary edge attention $\alpha_E$-(d) through the $\alpha_O$-(b) to cover the undetected areas. In refined object (e), we examined the regions, which were emphasized by $\alpha_O$-(b) and $\alpha_E$-(d).

\bibliography{ref}

\end{document}